\documentclass[10pt, a4paper]{article}
\usepackage{lrec}
\usepackage{multibib}
\newcites{languageresource}{Language Resources}
\usepackage{graphicx}
\usepackage{tabularx}
\usepackage{booktabs,siunitx}
\usepackage{svg}

\usepackage{soul}

\usepackage{epstopdf}
\usepackage[utf8]{inputenc}

\usepackage{hyperref}
\usepackage{xstring}

\usepackage{color}

\title{\textbf{An Annotated Corpus of Emerging Anglicisms in Spanish Newspaper Headlines}}

\name{Elena Álvarez-Mellado}

\address{Department of Computer Science, Brandeis University \\
         415 South St, Waltham, MA 02453 \\
         ealvarezmellado@brandeis.edu\\}

\abstract{
The extraction of anglicisms (lexical borrowings from English) is relevant both for lexicographic purposes and for NLP downstream tasks. We introduce a corpus of European Spanish newspaper headlines annotated with anglicisms and a baseline model for anglicism extraction. In this paper we present:  (1) a corpus of 21,570 newspaper headlines written in European Spanish annotated with emergent anglicisms and (2) a conditional random field baseline model with handcrafted features for anglicism extraction. We present the newspaper headlines corpus, describe the annotation tagset and guidelines and introduce a CRF model that can serve as baseline for the task of detecting anglicisms. The presented work is a first step towards the creation of an anglicism extractor for Spanish newswire. \\ \newline  }

\begin{document}

\maketitleabstract
\section{Introduction}
The study of English influence in the Spanish language has been a hot topic in Hispanic linguistics for decades, particularly concerning lexical borrowing or anglicisms \cite{capuz2004prestamos,lorenzo1996anglicismos,lopez1998anglicismo,menendez2003desplazamiento,nunez2017up,pratt1980anglicismo,gonzalez1999anglicisms}.

Lexical borrowing is a phenomenon that affects all languages and constitutes a productive mechanism for word-formation, especially in the press. \newcite{chesley_paula_predicting_2010} estimated that a reader of French newspapers encountered a new lexical borrowing for every 1,000 words. In Chilean newspapers, lexical borrowings account for approximately 30\% of neologisms, 80\% of those corresponding to English loanwords \cite{gerding2014anglicism}.

Detecting lexical borrowings is relevant both for lexicographic purposes and for NLP downstream tasks \cite{alex2007using,tsvetkov2016cross}. However, strategies to track and register lexical borrowings have traditionally relied on manual review of corpora. 

In this paper we present: (1) a corpus of newspaper headlines in European Spanish annotated with emerging anglicisms and (2) a CRF baseline model for anglicism automatic extraction in Spanish newswire.


\section{Related Work}
 Corpus-based studies of English borrowings in Spanish media have traditionally relied on manual evaluation of either  previously compiled general corpora such as CREA\footnote{\url{http://corpus.rae.es/creanet.html}} \cite{balteiro2011reassessment,nunez2016anglicisms,nogueroles2018corpus,oncins2012newly}, either new tailor-made corpora designed to analyze specific genres, varieties or phenomena \cite{de2012email,dieguez2004anglicismo,gerding2018neologia,nunez2017typographical,patzelt2011impact,rodriguez2002anglicismos,velez2003anglicismos}. 

In terms of automatic detection of anglicisms, previous approaches in different languages have mostly depended on resource lookup (lexicon or corpus frequencies), character n-grams and pattern matching. \newcite{alex-2008-comparing} combined lexicon lookup and a search engine module that used the web as a corpus to detect English inclusions in a corpus of German texts and compared her results with a maxent Markov model. \newcite{furiassi2007retrieval} explored corpora lookup and character n-grams to extract false anglicisms from a corpus of Italian newspapers. \newcite{andersen2012semi} used dictionary lookup, regular expressions and lexicon-derived frequencies of character n-grams to detect anglicism candidates in the Norwegian Newspaper Corpus (NNC) \cite{hofland-2000-self}, while \newcite{losnegaard2012data} explored a Machine Learning approach to anglicism detection in Norwegian by using TiMBL (Tilburg Memory-Based Learner, an implementation of a k-nearest neighbor classifier) with character trigrams as features. \newcite{garley-hockenmaier-2012-beefmoves} trained a maxent classifier
with character n-gram and morphological features
to identify anglicisms in German online communities. 

In Spanish, \newcite{serigos2017using} extracted anglicisms from a  corpus of Argentinian newspapers by combining dictionary lookup (aided by TreeTagger and the NLTK lemmatizer) with automatic filtering of capitalized words and manual inspection. In \newcite{serigos2017applying}, a character n-gram module was added to estimate the probabilities of a word being English or Spanish. \newcite{moreno2018configuracion} used different pattern-matching filters and lexicon lookup to extract anglicism cadidates from a corpus of tweets in US Spanish. 

Work within the code-switching community has also dealt with language identification on multilingual corpora. Due to the nature of code-switching, these models have primarily focused on oral copora and social media datasets \cite{aguilar-etal-2018-named,molina-etal-2016-overview,solorio-etal-2014-overview}. In the last shared task of language identification in code-switched data \cite{molina-etal-2016-overview}, approaches to English-Spanish included CRFs models  \cite{al-badrashiny-diab-2016-george,shrestha-2016-codeswitching,sikdar-gamback-2016-language,xia-2016-codeswitching}, logistic regression \cite{shirvani-etal-2016-howard} and LSTMs models \cite{jaech-etal-2016-neural,samih-etal-2016-multilingual}. 

The scope and nature of lexical borrowing is, however, somewhat different to that of code-switching. In fact, applying code-switching models to lexical borrowing detection has previously proved to be unsuccessful, as they tend to overestimate the number of anglicisms \cite{serigos2017applying}. In the next section we address the differences between both phenomena and set the scope of this project.       

\section{Anglicism: Scope of the Phenomenon}
Linguistic borrowing can be defined as the transference of linguistic elements between two languages. Borrowing  and code-switching have frequently been described as a continuum \cite{clyne2003dynamics}, with a fuzzy frontier between the two. As a result, a precise definition of what borrowing is remains elusive \cite{gomez1997towards} and some authors prefer to talk about code-mixing in general  \cite{alex2008automatic} or ``lone other-language incorporations" \cite{poplack2012myths}.   

Lexical borrowing in particular involves the incorporation of single lexical units from one language into another language and is usually accompanied by morphological and phonological modification to conform with the patterns of the recipient language \cite{onysko2007anglicisms,poplack1988social}. By definition, code-switches are not integrated into a recipient language, unlike
established loanwords \cite{poplack2012does}. While code-switches are usually fluent multiword interferences that normally comply with grammatical restrictions in both languages and that are produced by bilingual speakers in bilingual discourses, lexical borrowings are words used by monolingual individuals that eventually become lexicalized and assimilated as part of the recipient language lexicon until the knowledge of ``foreign" origin disappears \cite{lipski2005code}. 

In terms of approaching the problem, automatic code-switching identification has been framed as a sequence modeling problem where every token receives a language ID label (as in a POS-tagging task). Borrowing detection, on the other hand, while it can also be transformed into a sequence labeling problem, is an extraction task, where only certain spans of texts will be labeled (in the fashion of a NER task).

Various typologies have been proposed that aim to classify borrowings according to different criteria, both with a cross-linguistic perspective and also specifically aimed to characterize English inclusions in Spanish \cite{gomez1997towards,haspelmath2008loanword,nogueroles2018comprehensive,pratt1980anglicismo}. In this work, we will be focusing on unassimilated lexical borrowings (sometimes called \textit{foreignisms}), i.e. words from English origin that are introduced into  Spanish without any morphological or orthographic adaptation.  

\section{Corpus description and annotation}
\subsection{Corpus description}
In this subsection we describe the characteristics of the corpus. We first introduce the main corpus, with the usual train/development/test split that was used to train, tune and evaluate the model. We then present an additional test set that was designed to assess the performance of the model on more naturalistic data.
\subsubsection{Main Corpus}
The main corpus consists of a collection of monolingual newspaper headlines written in European Spanish. The corpus contains 16,553 headlines, which amounts to 244,114 tokens. Out of those 16,553 headlines, 1,109 contain at least one anglicism. The total number of anglicisms is 1,176 (most of them are a single word, although some of them were multiword expressions). The corpus was divided into training, development and test set. The proportions of headlines, tokens and anglicisms in each corpus split can be found in Table \ref{tab:split}.

The headlines in this corpus come from the Spanish newspaper \textit{eldiario.es}\footnote{\url{http://www.eldiario.es/}}, a progressive online newspaper based in Spain. \textit{eldiario.es} is one of the main national newspapers from Spain and, to the best of our knowledge, the only one that publishes its content under a Creative Commons license, which made it ideal for making the corpus publicly available\footnote{Both the corpus and the baseline model (Section 5) can be found at \url{https://github.com/lirondos/lazaro}.}.

\begin{table}[ht]
\centering
\resizebox{\columnwidth}{!}{\begin{tabular}[t]{lrrrrr}
\toprule
\textbf{\centering Set} & \textbf{\centering Headlines} & \textbf{\centering Tokens} & \textbf{\centering Headlines} & \textbf{\centering Anglicisms} & \textbf{\centering Other}\\
\newline  &  &  & \textbf{\centering with anglicisms} &  & \textbf{\centering borrowings}\\
\midrule
Train & 10,513 & 154,632 & 709 & 747 & 40\\
Dev & 3,020 & 44,758 & 200 & 219 & 14 \\
Test & 3,020 & 44,724 & 202 & 212 & 13\\
Suppl. test & 5,017 & 81,551 & 122 & 126 & 35\\
\bottomrule
\end{tabular}}
\caption{Number of headlines, tokens and anglicisms per corpus subset.}
\label{tab:split}
\end{table}%

The headlines were extracted from the newspaper website through web scraping  and range from September 2012 to January 2020. Only the following sections were included: economy, technology, lifestyle, music, TV and opinion. These sections were chosen as they were the most likely to contain anglicisms. The proportion of headlines with anglicisms per section can be found in Table \ref{tab:sections}. 

\begin{table}[ht]
\centering
\begin{tabular}[t]{lr}
\toprule
\textbf{Section} &\textbf{Percentage of anglicisms}\\
\midrule
      Opinion & 2.54\%\\
    Economy & 3.70\%\\
    Lifestyle & 6.48\%\\
      TV & 8.83\%\\
    Music & 9.25\%\\
    Technology & 15.37\%\\
\bottomrule
\end{tabular}
\caption{Percentage of headlines with anglicisms per section.}
\label{tab:sections}
\end{table}%

Using headlines (instead of full articles) was beneficial for several reasons. First of all, annotating a headline is faster and easier than annotating a full article; this helps ensure that a wider variety of topics will be covered in the corpus. Secondly, anglicisms are abundant in headlines, because they are frequently used as a way of calling the attention of the reader \cite{furiassi2007retrieval}. Finally, borrowings that make it to the headline are likely to be particularly salient or relevant, and therefore are good candidates for being extracted and tracked. 

\subsubsection{Supplemental Test Set}
In addition to the usual train/development/test split we have just presented, a supplemental test set of 5,017 headlines was collected. The headlines included in this additional test set also belong to \textit{eldiario.es}. These headlines were retrieved daily through RSS during February 2020 and included all sections from the newspaper. The headlines in the supplemental corpus therefore do not overlap in time with the main corpus and include more sections. The number of headlines, tokens and anglicisms in the supplemental test set can be found in Table \ref{tab:split}.

The motivation behind this supplemental test set is to assess the model performance on more naturalistic data, as the headlines in the supplemental corpus (1) belong to the future of the main corpus and (2) come from a less borrowing-dense sample. This supplemental test set better mimics the real scenario that an actual anglicism extractor would face and can be used to assess how well the model generalizes to detect anglicisms in any section of the daily news, which is ultimately the aim of this project. 

\subsection{Annotation guidelines}
The term \textit{anglicism} covers a wide range of linguistic phenomena. Following the typology proposed by \newcite{gomez1997towards}, we focused on direct, unadapted, emerging Anglicisms, i.e. lexical borrowings from the English language into Spanish that have recently been imported and that have still not been assimilated into Spanish. Other phenomena such as semantic calques, syntactic anglicisms, acronyms and proper names were considered beyond the scope of this annotation project. 

Lexical borrowings can be adapted (the spelling of the word is modified to comply with the phonological and orthographic patterns of the recipient language) or unadapted (the word preserves its original spelling). For this annotation task, adapted borrowings were ignored and only unadapted borrowings were annotated. Therefore, Spanish adaptations of anglicisms like \textit{fútbol} (from \textit{football}), \textit{mitin} (from \textit{meeting}) and such were not annotated as borrowings. Similarly, words derived from foreign lexemes that do not comply with Spanish orthotactics but that have been morphologically derived following the Spanish paradigm (\textit{hacktivista}, \textit{hackear}, \textit{shakespeariano}) were not annotated either. However, pseudo-anglicisms (words that are formed as if they were English, but do not exist in English, such as \textit{footing} or \textit{balconing}) were annotated.

Words that were not adapted but whose original spelling complies with graphophonological rules of Spanish (and are therefore unlikely to be ever adapted, such as  \textit{web}, \textit{internet}, \textit{fan}, \textit{club}, \textit{videoclip}) were annotated or not depending on how recent or emergent they were. After all, a word like \textit{club}, that has been around in Spanish language for centuries, cannot be considered emergent anymore and, for this project, would not be as interesting to retrieve as real emerging anglicisms. The notion of \textit{emergent} is, however, time-dependent and quite subjective: in order to determine which unadapted, graphophonologically acceptable borrowings were to be annotated, the online version of the \textit{Diccionario de la lengua española}\footnote{\url{https://dle.rae.es/}} \citelanguageresource{dle} was consulted. This dictionary is compiled by the  Royal Spanish Academy, a prescriptive institution on Spanish language. This decision was motivated by the fact that, if a borrowing was already registered by this dictionary (that has conservative approach to language change) and is considered assimilated (that is, the institution recommended no italics or quotation marks to write that word) then it could be inferred that the word was not emergent anymore.

Although the previous guidelines covered most cases, they proved insufficient. Some anglicisms were unadapted (they preserved their original spelling), unacceptable according to the Spanish graphophonological rules, and yet did not satisfy the condition of being emergent. That was the case of words like \textit{jazz} or \textit{whisky}, words that do not comply with Spanish graphophonological rules but that were imported decades ago, cannot be considered emergent anymore and are unlikely to ever be adapted into the Spanish spelling system. To adjudicate these examples on those cases, the criterion of pragmatic markedness proposed by \newcite{winter2012proposing} (that distinguishes between catachrestic and non-catachrestic borrowing) was applied: if a borrowing was not adapted (i.e. its form remained exactly as it came from English) but referred to a particular invention or innovation that came via the English language, that was not perceived as new anymore and that had never competed with a Spanish equivalent, then it was ignored. This criteria  proved to be extremely useful to deal with old unadapted anglicisms in the fields of music and food. Figure 1 summarizes the decision steps followed during the annotation process.

The corpus was annotated by a native speaker of Spanish using Doccano\footnote{\url{https://github.com/chakki-works/doccano}} \citelanguageresource{doccano}. The annotation tagset includes two labels: \texttt{ENG}, to annotate the English borrowings just described, and \texttt{OTHER}. This \texttt{OTHER} tag was used to tag lexical borrowings from languages other than English. After all, although English is today by far the most prevalent donor of borrowings, there are other languages that also provide new borrowings to Spanish. Furthermore, the tag \texttt{OTHER} allows to annotate borrowings such as \textit{première} or \textit{tempeh}, borrowings that etymologically do not come from English but that have entered the Spanish language via English influence, even when their spelling is very different to English borrowings. In general, we considered that having such a tag could also help assess how successful a classifier is detecting foreign borrowings in general in Spanish newswire (without having to create a label for every possible donor language, as the number of examples would be too sparse). In total, the training set contained 40 entities labeled as \texttt{OTHER}, the development set contained 14 and the test set contained 13. The supplemental test set contained 35 \texttt{OTHER} entities.

\begin{figure}[!h]
\begin{center}
\includegraphics[width=\columnwidth, keepaspectratio, scale=0.5]{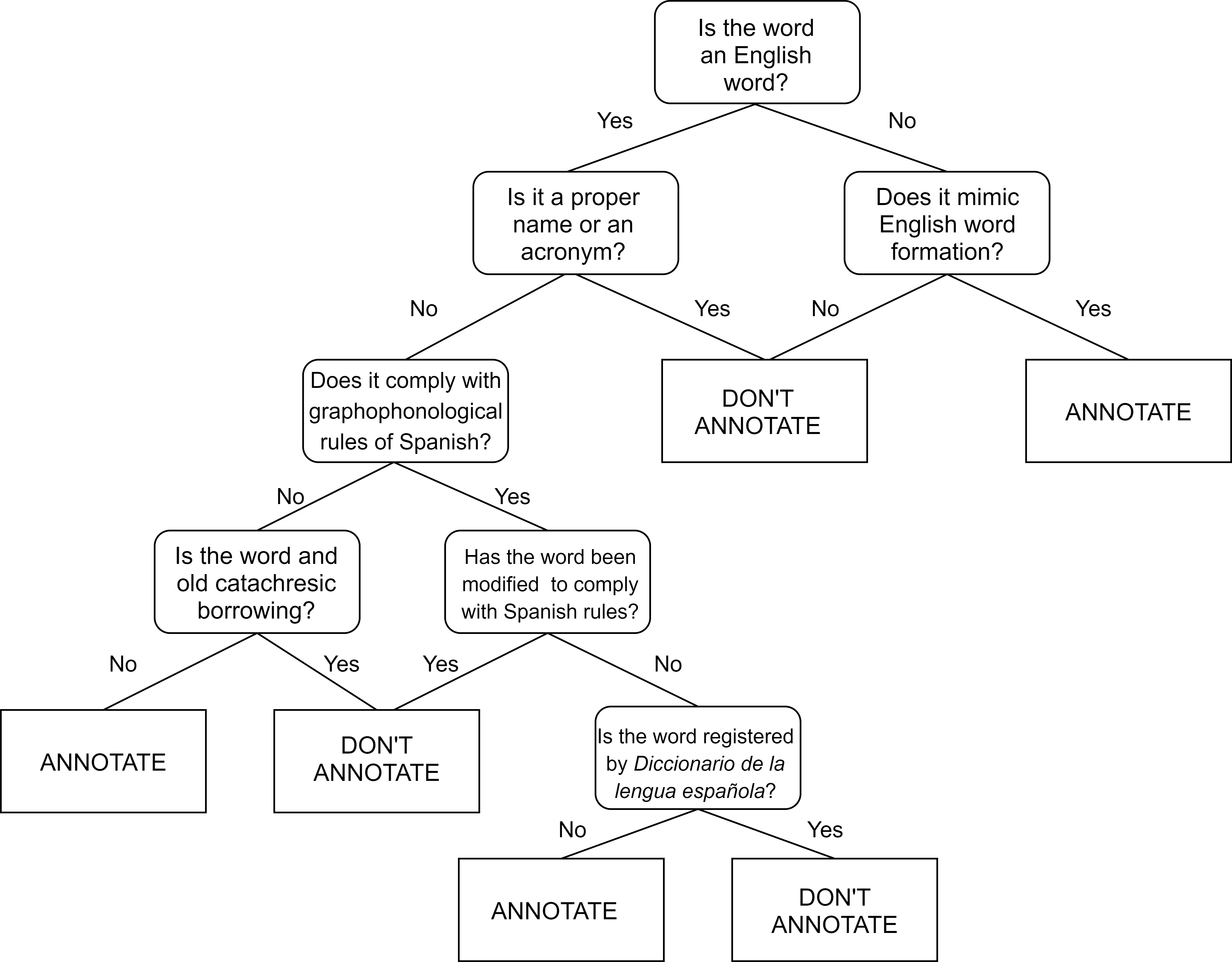} 
\caption{Decision steps to follow during the annotation process to decide whether to annotate a word as a borrowing.}
\label{fig1}
\end{center}
\end{figure}

\section{Baseline Model} \label{model}
A baseline model for automatic extraction of anglicisms was created using the annotated corpus we just presented as training material. As mentioned in Section 3, the task of detecting anglicisms can be approached as a sequence labeling problem where only certain spans of texts will be labeled as anglicism (in a similar way to an NER task). The chosen model was conditional random field model (CRF), which was also the most popular model in both Shared Tasks on Language Identification for Code-Switched Data \cite{molina-etal-2016-overview,solorio-etal-2014-overview}.

The model was built using \texttt{pycrfsuite}\footnote{\url{https://github.com/scrapinghub/python-crfsuite}}  \citelanguageresource{korobov2014python}, the Python wrapper for \texttt{crfsuite}\footnote{\url{ https://github.com/chokkan/crfsuite}} \citelanguageresource{CRFsuite} that implements CRF for labeling sequential data. It also used the \texttt{Token} and \texttt{Span} utilities from \texttt{spaCy}\footnote{\url{https://spacy.io/}} library \citelanguageresource{honnibal2017spacy}. 

The following handcrafted features were used for the model:
\begin{itemize}
    \item Bias feature
    \item Token feature
    \item Uppercase feature (y/n)
    \item Titlecase feature (y/n)
    \item Character trigram feature 
    \item Quotation feature (y/n)
    \item Word suffix feature (last three characters)
    \item POS tag (provided by \texttt{spaCy} utilities)
    \item Word shape (provided by \texttt{spaCy} utilities)
    \item Word embedding (see Table \ref{tab:embeddings})
\end{itemize}

Given that anglicisms can be multiword expressions (such as \textit{best seller, big data}) and that those units should be treated as one borrowing and not as two independent borrowings, we used multi-token BIO encoding to denote the boundaries of each span \cite{ramshaw1999text}. A window of two tokens in each direction was set for the feature extractor. The algorithm used was gradient descent with the L-BFGS method.

The model was tuned on the development set doing grid search; the hyperparameters considered were c1 (L1 regularization coefficient: $0.01$, $0.05$, $0.1$, $0.5$, $1.0$), c2 (L2 regularization coefficient: $0.01$, $0.05$, $0.1$, $0.5$, $1.0$), embedding scaling ($0.5$, $1.0$, $2.0$, $4.0$), and embedding type \citelanguageresource{bojanowski2017enriching,jose_canete_2019_3255001,cardellinoSBWCE,grave2018learning,honnibal2017spacy,perez_fasttext,perez_glove} (see Table \ref{tab:embeddings}). The best results were obtained with c1 = $0.05$, c2 = $0.01$, scaling = $0.5$ and word2vec Spanish embeddings by \newcite{cardellinoSBWCE}. The threshold for the stopping criterion delta was selected through observing the loss during preliminary experiments (delta = $1\mathrm{e}-3$). 

\begin{table}[ht]
\centering
\resizebox{\columnwidth}{!}{\begin{tabular}[t]{lcrr}
\toprule
\textbf{Author} & \textbf{Algorithm} & \textbf{\# Vectors} & \textbf{Dimensions}\\
\midrule
\newcite{bojanowski2017enriching} & FastText & 985,667 & 300\\
\newcite{jose_canete_2019_3255001} & FastText & 1,313,423 & 300\\
\newcite{cardellinoSBWCE} & word2vec & 1,000,653 & 300\\
\newcite{grave2018learning} & FastText & 2,000,001  & 300\\
\newcite{honnibal2017spacy} & word2vec & 534,000 & 50\\
\newcite{perez_fasttext} & FastText & 855,380 & 300\\
\newcite{perez_glove} & GloVe & 855,380 & 300\\
\bottomrule
\end{tabular}}
\caption{Types of embeddings tried.}
\label{tab:embeddings}
\end{table}%

In order to assess the significance of the the handcrafted features, a feature ablation study was done on the tuned  model, ablating one feature at a time and testing on the development set. Due to the scarcity of spans labeled with the \texttt{OTHER} tag on the development set (only 14) and given that the main purpose of the model is to detect anglicisms, the baseline model was run ignoring the \texttt{OTHER} tag both during tuning and the feature ablation experiments. Table~\ref{tab:ablation} displays the results on the development set with all features and for the different feature ablation runs. The results show that all features proposed for the baseline model contribute to the results, with the character trigram feature being the one that has the biggest impact on the feature ablation study.  

\begin{table}[ht]
\centering
\resizebox{\columnwidth}{!}{\begin{tabular}[t]{lrrrr}
\toprule
\textbf{Features} & \textbf{Precision} & \textbf{Recall} & \textbf{F1 score} & \textbf{F1 change}\\
\midrule
All features & 97.84 & \bf 82.65 & \bf 89.60 &  \\
$-$ Bias & 96.76 & 81.74 & 88.61 & $-$0.99\\
$-$ Token & 95.16 & 80.82 & 87.41 & $-$2.19\\
$-$ Uppercase & 97.30 & 82.19 & 89.11 & $-$0.49\\
$-$ Titlecase & 96.79 & \bf 82.65 & 89.16 & $-$0.44\\
$-$ Char trigram & 96.05 & 77.63 & 85.86 & $-$3.74\\
$-$ Quotation & 97.31 & \bf 82.65 & 89.38 & $-$0.22\\
$-$ Suffix & 97.30 & 82.19 & 89.11 & $-$0.49\\
$-$ POS tag & \bf 98.35 & 81.74 & 89.28 & $-$0.32\\
$-$ Word shape & 96.79 & \bf 82.65 & 89.16 & $-$0.44\\
$-$ Word embedding & 95.68 & 80.82 & 87.62 & $-$1.98\\
\bottomrule
\end{tabular}}
\caption{Ablation study results on the development test.}
\label{tab:ablation}
\end{table}%

\section{Results}
The baseline model was then run on the test set and the supplemental test set with the set of features and hyperparameters mentioned on Section \ref{model} Table \ref{tab:results} displays the results obtained. The model was run both with and without the \texttt{OTHER} tag. The metrics for \texttt{ENG} display the results obtained only for the spans labeled as anglicisms; the metrics for \texttt{OTHER} display the results obtained for any borrowing other than anglicisms. The metrics for \texttt{BORROWING} discard the type of label and consider correct any labeled span that has correct boundaries, regardless of the label type (so any type of borrowing, regardless if it is \texttt{ENG} or \texttt{OTHER}). In all cases, only full matches were considered correct and no credit was given to partial matching, i.e. if only \textit{fake} in \textit{fake news} was retrieved, it was considered wrong and no partial score was given. 

Results on all sets show an important difference between precision and recall, precision being significantly higher than recall. There is also a significant difference between the results obtained on development and test set (F1 = 89.60, F1 = 87.82) and the results on the supplemental test set (F1 = 71.49). The time difference between the supplemental test set and the development and test set (the headlines from the the supplemental test set being from a different time period to the training set) can probably explain these differences.

Comparing the results with and without the \texttt{OTHER} tag, it seems that including it on the development and test set produces worse results (or they remain roughly the same, at best). However, the best precision result on the supplemental test was obtained when including the \texttt{OTHER} tag and considering both \texttt{ENG} and \texttt{OTHER} spans as \texttt{BORROWING} (precision = 87.62). This is caused by the fact that, while the development and test set were compiled from anglicism-rich newspaper sections (similar to the training set), the supplemental test set contained headlines from all the sections in the newspaper, and therefore included borrowings from other languages such as Catalan, Basque or French. When running the model without the \texttt{OTHER} tag on the supplemental test set, these non-English borrowings were labeled as anglicisms by the model (after all, their spelling does not resemble Spanish spelling), damaging the precision score. When the \texttt{OTHER} tag was included, these non-English borrowings got correctly labeled as \texttt{OTHER}, improving the precision score. This proves that, although the \texttt{OTHER} tag might be irrelevant or even damaging when testing on the development or test set, it can be useful when testing on more naturalistic data, such as the one in the supplemental test set.

Concerning errors, two types of errors were recurrent among all sets: long titles of songs, films or series written in English were a source of false positives, as the model tended to mistake some of the uncapitalized words in the title for anglicisms (for example, \textit{it darker} in ```You want it darker', la oscura y brillante despedida de Leonard Cohen"). On the other hand, anglicisms that appear on the first position of the sentence (and were, therefore, capitalized) were consistently ignored (as the model probably assumed they were named entities) and produced a high number of false negatives (for example, \textit{vamping} in ``Vamping: la recurrente leyenda urbana de la luz azul `asesina'").   

\begin{table}[ht]
\centering
\resizebox{\columnwidth}{!}{\begin{tabular}[t]{lrrr}
\toprule
\textbf{Set} & \textbf{Precision} & \textbf{Recall} & \textbf{F1 score}\\
\midrule
Development set ($-$ \texttt{OTHER}) & 97.84 & 82.65 & 89.60\\
Development set ($+$ \texttt{OTHER}) &  &  & \\
\hspace{0.5cm} \texttt{ENG} & 96.79 & 82.65 & 89.16\\
\hspace{0.5cm} \texttt{OTHER} & 100.0 & 28.57 & 44.44\\
\hspace{0.5cm} \texttt{BORROWING} & 96.86 & 79.40 & 87.26\\\\

Test set ($-$ \texttt{OTHER}) & 95.05 & 81.60 & 87.82\\
Test set ($+$ \texttt{OTHER}) &  &  & \\
\hspace{0.5cm} \texttt{ENG} & 95.03 & 81.13 & 87.53\\
\hspace{0.5cm} \texttt{OTHER} & 100.0 & 46.15 & 63.16\\
\hspace{0.5cm} \texttt{BORROWING} & 95.19 & 79.11 & 86.41\\\\
Supplemental test set ($-$ \texttt{OTHER}) & 83.16  & 62.70 & 71.49 \\
Supplemental test set ($+$ \texttt{OTHER}) &  &  & \\
\hspace{0.5cm} \texttt{ENG} & 82.65 & 64.29 & 72.32\\
\hspace{0.5cm} \texttt{OTHER} & 100.0 & 20.0 & 33.33\\
\hspace{0.5cm} \texttt{BORROWING} & 87.62 & 57.14 & 69.17\\
\bottomrule
\end{tabular}}
\caption{Results on test set and supplemental test set.}
\label{tab:results}
\end{table}%



The results on Table \ref{tab:results} cannot, however, be compared to the ones reported by previous work: the metric that we report is span F-measure, as the evaluation was done on span level (instead of token level) and credit was only given to full matches. Secondly, there was no Spanish tag assigned to non-borrowings, that means that no credit was given if a Spanish token was identified as such.

\section{Future Work}
This is an on-going project. The corpus we have just presented is a first step towards the development of an extractor of emerging anglicisms in the Spanish press. Future work includes: assessing whether to keep the \texttt{OTHER} tag, improving the baseline model (particularly to improve recall), assessing the suitability and contribution of different sets of features and exploring different models. In terms of the corpus development, the training set is now closed and stable, but the test set could potentially be increased in order to have more and more diverse anglicisms. 

\section{Conclusions}
In this paper we have presented a new corpus of 21,570 newspaper headlines written in European Spanish. The corpus is annotated with emergent anglicisms and, up to our very best knowledge, is the first corpus of this type to be released publicly. We have presented the annotation scope, tagset and guidelines, and we have introduced a CRF baseline model for anglicism extraction trained with the described corpus. The results obtained show that the the corpus and baseline model are appropriate for automatic anglicism extraction.

\section{Acknowledgements}The author would like to thank Constantine Lignos for his feedback and advice on this project.

\section{Bibliographical References}\label{reference}

\bibliographystyle{lrec}

\begin{thebibliography}{}

\bibitem[\protect\citename{Aguilar \bgroup et al.\egroup
  }2018]{aguilar-etal-2018-named}
Aguilar, G., AlGhamdi, F., Soto, V., Diab, M., Hirschberg, J., and Solorio, T.
\newblock (2018).
\newblock Named entity recognition on code-switched data: Overview of the
  {CALCS} 2018 shared task.
\newblock In {\em Proceedings of the Third Workshop on Computational Approaches
  to Linguistic Code-Switching}, pages 138--147, Melbourne, Australia, July.
  Association for Computational Linguistics.

\bibitem[\protect\citename{Al-Badrashiny and
  Diab}2016]{al-badrashiny-diab-2016-george}
Al-Badrashiny, M. and Diab, M.
\newblock (2016).
\newblock The {G}eorge {W}ashington {U}niversity {S}ystem for the
  {C}ode-{S}witching {W}orkshop {S}hared {T}ask 2016.
\newblock In {\em Proceedings of the Second Workshop on Computational
  Approaches to Code Switching}, pages 108--111, Austin, Texas, November.
  Association for Computational Linguistics.

\bibitem[\protect\citename{Alex \bgroup et al.\egroup }2007]{alex2007using}
Alex, B., Dubey, A., and Keller, F.
\newblock (2007).
\newblock Using foreign inclusion detection to improve parsing performance.
\newblock In {\em Proceedings of the 2007 Joint Conference on Empirical Methods
  in Natural Language Processing and Computational Natural Language Learning
  (EMNLP-CoNLL)}, pages 151--160.

\bibitem[\protect\citename{Alex}2008a]{alex2008automatic}
Alex, B.
\newblock (2008a).
\newblock {\em Automatic detection of English inclusions in mixed-lingual data
  with an application to parsing}.
\newblock {Ph.D.} thesis, University of Edinburgh.

\bibitem[\protect\citename{Alex}2008b]{alex-2008-comparing}
Alex, B.
\newblock (2008b).
\newblock Comparing corpus-based to web-based lookup techniques for automatic
  {E}nglish inclusion detection.
\newblock In {\em Proceedings of the Sixth International Conference on Language
  Resources and Evaluation ({LREC}'08)}, Marrakech, Morocco, May. European
  Language Resources Association (ELRA).

\bibitem[\protect\citename{Andersen}2012]{andersen2012semi}
Andersen, G.
\newblock (2012).
\newblock Semi-automatic approaches to {A}nglicism detection in {N}orwegian
  corpus data.
\newblock In Cristiano Furiassi, et~al., editors, {\em The anglicization of
  European lexis}, pages 111--130.

\bibitem[\protect\citename{Balteiro}2011]{balteiro2011reassessment}
Balteiro, I.
\newblock (2011).
\newblock A reassessment of traditional lexicographical tools in the light of
  new corpora: sports anglicisms in {S}panish.
\newblock {\em International Journal of English Studies}, 11(2):23--52.

\bibitem[\protect\citename{Bojanowski \bgroup et al.\egroup
  }2017]{bojanowski2017enriching}
Bojanowski, P., Grave, E., Joulin, A., and Mikolov, T.
\newblock (2017).
\newblock Enriching word vectors with subword information.
\newblock {\em Transactions of the Association for Computational Linguistics},
  5:135--146.

\bibitem[\protect\citename{Chesley and
  Baayen}2010]{chesley_paula_predicting_2010}
Chesley, P. and Baayen, R.~H.
\newblock (2010).
\newblock Predicting new words from newer words: {Lexical} borrowings in
  {French}.
\newblock {\em Linguistics}, 48(6):1343.

\bibitem[\protect\citename{Clyne \bgroup et al.\egroup
  }2003]{clyne2003dynamics}
Clyne, M., Clyne, M.~G., and Michael, C.
\newblock (2003).
\newblock {\em Dynamics of language contact: English and immigrant languages}.
\newblock Cambridge University Press.

\bibitem[\protect\citename{De~la Cruz~Cabanillas and
  Mart{\'\i}nez}2012]{de2012email}
De~la Cruz~Cabanillas, I. and Mart{\'\i}nez, C.~T.
\newblock (2012).
\newblock Email or correo electr{\'o}nico? {A}nglicisms in {S}panish.
\newblock {\em Revista espa{\~n}ola de ling{\"u}{\'\i}stica aplicada},
  (1):95--118.

\bibitem[\protect\citename{Di{\'e}guez}2004]{dieguez2004anglicismo}
Di{\'e}guez, M.~I.
\newblock (2004).
\newblock El anglicismo l{\'e}xico en el discurso econ{\'o}mico de
  divulgaci{\'o}n cient{\'\i}fica del espa{\~n}ol de {C}hile.
\newblock {\em Onom{\'a}zein}, 2(10):117--141.

\bibitem[\protect\citename{Furiassi and Hofland}2007]{furiassi2007retrieval}
Furiassi, C. and Hofland, K.
\newblock (2007).
\newblock The retrieval of false anglicisms in newspaper texts.
\newblock In {\em Corpus Linguistics 25 Years On}, pages 347--363. Brill
  Rodopi.

\bibitem[\protect\citename{Garley and
  Hockenmaier}2012]{garley-hockenmaier-2012-beefmoves}
Garley, M. and Hockenmaier, J.
\newblock (2012).
\newblock {B}eefmoves: Dissemination, diversity, and dynamics of {E}nglish
  borrowings in a {G}erman hip hop forum.
\newblock In {\em Proceedings of the 50th Annual Meeting of the Association for
  Computational Linguistics (Volume 2: Short Papers)}, pages 135--139, Jeju
  Island, Korea, July. Association for Computational Linguistics.

\bibitem[\protect\citename{Gerding \bgroup et al.\egroup
  }2014]{gerding2014anglicism}
Gerding, C., Fuentes, M., G{\'o}mez, L., and Kotz, G.
\newblock (2014).
\newblock Anglicism: An active word-formation mechanism in {S}panish.
\newblock {\em Colombian Applied Linguistics Journal}, 16(1):40--54.

\bibitem[\protect\citename{Gerding~Salas \bgroup et al.\egroup
  }2018]{gerding2018neologia}
Gerding~Salas, C., Ca{\~n}ete~Gonz{\'a}lez, P., and Adam, C.
\newblock (2018).
\newblock Neolog{\'\i}a sintagm{\'a}tica anglicada en espa{\~n}ol: Calcos y
  pr{\'e}stamos.
\newblock {\em Revista signos}, 51(97):175--192.

\bibitem[\protect\citename{G{\'o}mez~Capuz}1997]{gomez1997towards}
G{\'o}mez~Capuz, J.
\newblock (1997).
\newblock Towards a typological classification of linguistic borrowing
  (illustrated with anglicisms in romance languages).
\newblock {\em Revista alicantina de estudios ingleses}, 10:81--94.

\bibitem[\protect\citename{Grave \bgroup et al.\egroup
  }2018]{grave2018learning}
Grave, E., Bojanowski, P., Gupta, P., Joulin, A., and Mikolov, T.
\newblock (2018).
\newblock Learning word vectors for 157 languages.
\newblock In {\em Proceedings of the International Conference on Language
  Resources and Evaluation (LREC 2018)}.

\bibitem[\protect\citename{Gómez~Capuz}2004]{capuz2004prestamos}
Gómez~Capuz, J.
\newblock (2004).
\newblock {\em Los pr{\'e}stamos del espa{\~n}ol: lengua y sociedad}.
\newblock Cuadernos de Lengua Espa{\'n}ola. Arco Libros.

\bibitem[\protect\citename{Haspelmath}2008]{haspelmath2008loanword}
Haspelmath, M.
\newblock (2008).
\newblock Loanword typology: Steps toward a systematic cross-linguistic study
  of lexical borrowability.
\newblock {\em Empirical Approaches to Language Typology}, 35:43.

\bibitem[\protect\citename{Hofland}2000]{hofland-2000-self}
Hofland, K.
\newblock (2000).
\newblock A self-expanding corpus based on newspapers on the web.
\newblock In {\em Proceedings of the Second International Conference on
  Language Resources and Evaluation ({LREC}{'}00)}, Athens, Greece, May.
  European Language Resources Association (ELRA).

\bibitem[\protect\citename{Jaech \bgroup et al.\egroup
  }2016]{jaech-etal-2016-neural}
Jaech, A., Mulcaire, G., Ostendorf, M., and Smith, N.~A.
\newblock (2016).
\newblock A neural model for language identification in code-switched tweets.
\newblock In {\em Proceedings of the Second Workshop on Computational
  Approaches to Code Switching}, pages 60--64, Austin, Texas, November.
  Association for Computational Linguistics.

\bibitem[\protect\citename{Lipski}2005]{lipski2005code}
Lipski, J.~M.
\newblock (2005).
\newblock Code-switching or borrowing? {N}o s{\'e} so no puedo decir, you know.
\newblock In {\em Selected proceedings of the second workshop on Spanish
  sociolinguistics}, pages 1--15. Cascadilla Proceedings Project Somerville,
  MA.

\bibitem[\protect\citename{Lorenzo}1996]{lorenzo1996anglicismos}
Lorenzo, E.
\newblock (1996).
\newblock {\em Anglicismos hisp{\'a}nicos}.
\newblock Biblioteca rom{\'a}nica hisp{\'a}nica: Estudios y ensayos. Gredos.

\bibitem[\protect\citename{Losnegaard and Lyse}2012]{losnegaard2012data}
Losnegaard, G.~S. and Lyse, G.~I.
\newblock (2012).
\newblock A data-driven approach to anglicism identification in {N}orwegian.
\newblock In Gisle Andersen, editor, {\em Exploring Newspaper Language: Using
  the web to create and investigate a large corpus of modern Norwegian}, pages
  131--154. John Benjamins Publishing.

\bibitem[\protect\citename{Medina~L{\'o}pez}1998]{lopez1998anglicismo}
Medina~L{\'o}pez, J.
\newblock (1998).
\newblock {\em El anglicismo en el espa{\~n}ol actual}.
\newblock Cuadernos de lengua espa{\~n}ola. Arco Libros.

\bibitem[\protect\citename{Men{\'e}ndez \bgroup et al.\egroup
  }2003]{menendez2003desplazamiento}
Men{\'e}ndez, F., Men{\'e}ndez, M., and Morales, H.
\newblock (2003).
\newblock {\em El desplazamiento ling{\"u}{\'\i}stico del espa{\~n}ol por el
  ingl{\'e}s}.
\newblock C{\'a}tedra ling{\"u}{\'\i}stica. C{\'a}tedra.

\bibitem[\protect\citename{Molina \bgroup et al.\egroup
  }2016]{molina-etal-2016-overview}
Molina, G., AlGhamdi, F., Ghoneim, M., Hawwari, A., Rey-Villamizar, N., Diab,
  M., and Solorio, T.
\newblock (2016).
\newblock Overview for the second shared task on language identification in
  code-switched data.
\newblock In {\em Proceedings of the Second Workshop on Computational
  Approaches to Code Switching}, pages 40--49, Austin, Texas, November.
  Association for Computational Linguistics.

\bibitem[\protect\citename{Moreno~Fern{\'a}ndez and
  Moreno~Sandoval}2018]{moreno2018configuracion}
Moreno~Fern{\'a}ndez, F. and Moreno~Sandoval, A.
\newblock (2018).
\newblock Configuraci{\'o}n ling{\"u}{\'\i}stica de anglicismos procedentes de
  {T}witter en el espa{\~n}ol estadounidense.
\newblock {\em Revista signos}, 51(98):382--409.

\bibitem[\protect\citename{N{\'u}{\~n}ez~Nogueroles}2016]{nunez2016anglicisms}
N{\'u}{\~n}ez~Nogueroles, E.~E.
\newblock (2016).
\newblock Anglicisms in {CREA}: a quantitative analysis in {S}panish
  newspapers.
\newblock {\em Language design: journal of theoretical and experimental
  linguistics}, 18:0215--242.

\bibitem[\protect\citename{N{\'u}{\~n}ez~Nogueroles}2017a]{nunez2017up}
N{\'u}{\~n}ez~Nogueroles, E.
\newblock (2017a).
\newblock An up-to-date review of the literature on anglicisms in spanish.
\newblock {\em Di{\'a}logo de la Lengua, IX}, pages 1--54.

\bibitem[\protect\citename{N{\'u}{\~n}ez~Nogueroles}2017b]{nunez2017typographical}
N{\'u}{\~n}ez~Nogueroles, E.~E.
\newblock (2017b).
\newblock Typographical, orthographic and morphological variation of anglicisms
  in a corpus of {S}panish newspaper texts.
\newblock {\em Revista Canaria de Estudios Ingleses}, (75):175--190.

\bibitem[\protect\citename{N{\'u}{\~n}ez~Nogueroles}2018a]{nogueroles2018comprehensive}
N{\'u}{\~n}ez~Nogueroles, E.~E.
\newblock (2018a).
\newblock A comprehensive definition and typology of anglicisms in present-day
  {S}panish.
\newblock {\em Epos: Revista de filolog{\'\i}a}, (34):211--237.

\bibitem[\protect\citename{N{\'u}{\~n}ez~Nogueroles}2018b]{nogueroles2018corpus}
N{\'u}{\~n}ez~Nogueroles, E.~E.
\newblock (2018b).
\newblock A corpus-based study of anglicisms in the 21st century spanish press.
\newblock {\em Analecta Malacitana (AnMal electr{\'o}nica)}, (44):123--159.

\bibitem[\protect\citename{{O}nc{\'\i}ns {M}art{\'\i}nez}2012]{oncins2012newly}
{O}nc{\'\i}ns {M}art{\'\i}nez, J.~L.
\newblock (2012).
\newblock Newly-coined anglicisms in contemporary {S}panish. a corpus-based
  approach.
\newblock In Cristiano Furiassi, et~al., editors, {\em The anglicization of
  European lexis}, pages 217--238.

\bibitem[\protect\citename{Onysko}2007]{onysko2007anglicisms}
Onysko, A.
\newblock (2007).
\newblock {\em Anglicisms in German: Borrowing, lexical productivity, and
  written codeswitching}, volume~23.
\newblock Walter de Gruyter.

\bibitem[\protect\citename{Patzelt}2011]{patzelt2011impact}
Patzelt, C.
\newblock (2011).
\newblock The impact of {E}nglish on {S}panish-language media in the usa.
\newblock In {\em Multilingual Discourse Production: Diachronic and Synchronic
  Perspectives}, volume~12, page 257. John Benjamins Publishing.

\bibitem[\protect\citename{Poplack and Dion}2012]{poplack2012myths}
Poplack, S. and Dion, N.
\newblock (2012).
\newblock Myths and facts about loanword development.
\newblock {\em Language Variation and Change}, 24(3):279--315.

\bibitem[\protect\citename{Poplack \bgroup et al.\egroup
  }1988]{poplack1988social}
Poplack, S., Sankoff, D., and Miller, C.
\newblock (1988).
\newblock The social correlates and linguistic processes of lexical borrowing
  and assimilation.
\newblock {\em Linguistics}, 26(1):47--104.

\bibitem[\protect\citename{Poplack}2012]{poplack2012does}
Poplack, S.
\newblock (2012).
\newblock What does the nonce borrowing hypothesis hypothesize?
\newblock {\em Bilingualism: Language and Cognition}, 15(3):644--648.

\bibitem[\protect\citename{Pratt}1980]{pratt1980anglicismo}
Pratt, C.
\newblock (1980).
\newblock {\em El anglicismo en el espa{\~n}ol peninsular contempor{\'a}neo},
  volume 308.
\newblock Gredos.

\bibitem[\protect\citename{Ramshaw and Marcus}1999]{ramshaw1999text}
Ramshaw, L.~A. and Marcus, M.~P.
\newblock (1999).
\newblock Text chunking using transformation-based learning.
\newblock In {\em Natural language processing using very large corpora}, pages
  157--176. Springer.

\bibitem[\protect\citename{Rodr{\'\i}guez~Medina}2002]{rodriguez2002anglicismos}
Rodr{\'\i}guez~Medina, M.~J.
\newblock (2002).
\newblock Los anglicismos de frecuencia sint{\'a}cticos en espa{\~n}ol: estudio
  emp{\'\i}rico.
\newblock {\em RAEL. Revista electr{\'o}nica de ling{\"u}{\'\i}stica aplicada}.

\bibitem[\protect\citename{Rodríguez~González}1999]{gonzalez1999anglicisms}
Rodríguez~González, F.
\newblock (1999).
\newblock Anglicisms in contemporary {S}panish. an overview.
\newblock {\em Atlantis}, 21(1/2):103--139.

\bibitem[\protect\citename{Samih \bgroup et al.\egroup
  }2016]{samih-etal-2016-multilingual}
Samih, Y., Maharjan, S., Attia, M., Kallmeyer, L., and Solorio, T.
\newblock (2016).
\newblock Multilingual code-switching identification via {LSTM} recurrent
  neural networks.
\newblock In {\em Proceedings of the Second Workshop on Computational
  Approaches to Code Switching}, pages 50--59, Austin, Texas, November.
  Association for Computational Linguistics.

\bibitem[\protect\citename{Serigos}2017a]{serigos2017using}
Serigos, J.
\newblock (2017a).
\newblock Using distributional semantics in loanword research: A concept-based
  approach to quantifying semantic specificity of anglicisms in {S}panish.
\newblock {\em International Journal of Bilingualism}, 21(5):521--540.

\bibitem[\protect\citename{Serigos}2017b]{serigos2017applying}
Serigos, J. R.~L.
\newblock (2017b).
\newblock {\em Applying corpus and computational methods to loanword research:
  new approaches to Anglicisms in {S}panish}.
\newblock {Ph.D.} thesis, The University of Texas at Austin.

\bibitem[\protect\citename{Shirvani \bgroup et al.\egroup
  }2016]{shirvani-etal-2016-howard}
Shirvani, R., Piergallini, M., Gautam, G.~S., and Chouikha, M.
\newblock (2016).
\newblock The {H}oward {U}niversity {S}ystem submission for the {S}hared {T}ask
  in {L}anguage {I}dentification in {S}panish-{E}nglish {C}odeswitching.
\newblock In {\em Proceedings of the Second Workshop on Computational
  Approaches to Code Switching}, pages 116--120, Austin, Texas, November.
  Association for Computational Linguistics.

\bibitem[\protect\citename{Shrestha}2016]{shrestha-2016-codeswitching}
Shrestha, P.
\newblock (2016).
\newblock Codeswitching detection via lexical features in conditional random
  fields.
\newblock In {\em Proceedings of the Second Workshop on Computational
  Approaches to Code Switching}, pages 121--126, Austin, Texas, November.
  Association for Computational Linguistics.

\bibitem[\protect\citename{Sikdar and
  Gamb{\"a}ck}2016]{sikdar-gamback-2016-language}
Sikdar, U.~K. and Gamb{\"a}ck, B.
\newblock (2016).
\newblock Language identification in code-switched text using conditional
  random fields and {B}abelnet.
\newblock In {\em Proceedings of the Second Workshop on Computational
  Approaches to Code Switching}, pages 127--131, Austin, Texas, November.
  Association for Computational Linguistics.

\bibitem[\protect\citename{Solorio \bgroup et al.\egroup
  }2014]{solorio-etal-2014-overview}
Solorio, T., Blair, E., Maharjan, S., Bethard, S., Diab, M., Ghoneim, M.,
  Hawwari, A., AlGhamdi, F., Hirschberg, J., Chang, A., and Fung, P.
\newblock (2014).
\newblock Overview for the first shared task on language identification in
  code-switched data.
\newblock In {\em Proceedings of the First Workshop on Computational Approaches
  to Code Switching}, pages 62--72, Doha, Qatar, October. Association for
  Computational Linguistics.

\bibitem[\protect\citename{Tsvetkov and Dyer}2016]{tsvetkov2016cross}
Tsvetkov, Y. and Dyer, C.
\newblock (2016).
\newblock Cross-lingual bridges with models of lexical borrowing.
\newblock {\em Journal of Artificial Intelligence Research}, 55:63--93.

\bibitem[\protect\citename{V{\'e}lez~Barreiro}2003]{velez2003anglicismos}
V{\'e}lez~Barreiro, M.
\newblock (2003).
\newblock {\em Anglicismos en la prensa econ{\'o}mica espa{\~n}ola}.
\newblock {Ph.D.} thesis, Universidade da Coruña.

\bibitem[\protect\citename{Winter-Froemel and Onysko}2012]{winter2012proposing}
Winter-Froemel, E. and Onysko, A.
\newblock (2012).
\newblock Proposing a pragmatic distinction for lexical anglicisms.
\newblock In Cristiano Furiassi, et~al., editors, {\em The anglicization of
  European lexis}, page~43.

\bibitem[\protect\citename{Xia}2016]{xia-2016-codeswitching}
Xia, M.~X.
\newblock (2016).
\newblock Codeswitching language identification using subword information
  enriched word vectors.
\newblock In {\em Proceedings of the Second Workshop on Computational
  Approaches to Code Switching}, pages 132--136, Austin, Texas, November.
  Association for Computational Linguistics.

\end{thebibliography}

\begin{thebibliography}{}

\bibitem[\protect\citename{Cardellino}2019]{cardellinoSBWCE}
Cardellino, Cristian.
\newblock (2019).
\newblock {\em Spanish {B}illion {W}ords {C}orpus and {E}mbeddings}.
\newblock \url{https://crscardellino.github.io/SBWCE/}.

\bibitem[\protect\citename{Cañete}2019]{jose_canete_2019_3255001}
Cañete, José.
\newblock (2019).
\newblock {\em Spanish Word Embeddings}.
\newblock Zenodo, \url{https://doi.org/10.5281/zenodo.3255001}.

\bibitem[\protect\citename{Honnibal and Montani}2017]{honnibal2017spacy}
Honnibal, Matthew and Montani, Ines.
\newblock (2017).
\newblock {\em spa{C}y 2: Natural language understanding with bloom embeddings,
  convolutional neural networks and incremental parsing}.
\newblock \url{https://spacy.io/}.

\bibitem[\protect\citename{Korobov and Peng}2014]{korobov2014python}
Korobov, M and Peng, T.
\newblock (2014).
\newblock {\em Python-crfsuite}.
\newblock \url{https://github.com/scrapinghub/python-crfsuite}.

\bibitem[\protect\citename{Nakayama \bgroup et al.\egroup }2018]{doccano}
Hiroki Nakayama and Takahiro Kubo and Junya Kamura and Yasufumi Taniguchi and
  Xu Liang.
\newblock (2018).
\newblock {\em {doccano}: Text Annotation Tool for Human}.
\newblock \url{https://github.com/doccano/doccano}.

\bibitem[\protect\citename{Okazaki}2007]{CRFsuite}
Naoaki Okazaki.
\newblock (2007).
\newblock {\em CRFsuite: a fast implementation of {C}onditional {R}andom
  {F}ields ({C}{R}{F}s)}.
\newblock \url{http://www.chokkan.org/software/crfsuite/}.

\bibitem[\protect\citename{Pérez}2017a]{perez_fasttext}
Pérez, Jorge.
\newblock (2017a).
\newblock {\em FastText embeddings from SBWC}.
\newblock Available at
  \url{https://github.com/dccuchile/spanish-word-embeddings#fasttext-embeddings-from-sbwc}.

\bibitem[\protect\citename{Pérez}2017b]{perez_glove}
Pérez, Jorge.
\newblock (2017b).
\newblock {\em GloVe embeddings from SBWC}.
\newblock Available at
  \url{https://github.com/dccuchile/spanish-word-embeddings#glove-embeddings-from-sbwc}.

\bibitem[\protect\citename{{Real Academia Espa{\~n}ola}}2014]{dle}
{Real Academia Espa{\~n}ola}.
\newblock (2014).
\newblock {\em Diccionario de la lengua espa{\~n}ola, ed. 23.3}.
\newblock \url{http://dle. rae. es}.

\end{thebibliography}

\section{Language Resource References}
\label{lr:ref}
\bibliographystylelanguageresource{lrec}

\end{document}